\documentclass[letterpaper]{article} 
\usepackage{aaai2026}  
\usepackage{times}  
\usepackage{helvet}  
\usepackage{courier}  
\usepackage[hyphens]{url}  
\usepackage{graphicx} 
\usepackage{natbib}  
\usepackage{caption} 
\usepackage{algorithm}
\usepackage{algorithmic}

\usepackage{newfloat}
\usepackage{listings}
\usepackage{booktabs}
\usepackage{subcaption}
\usepackage{amsmath,amssymb}
\usepackage{listings}
\DeclareCaptionStyle{ruled}{labelfont=normalfont,labelsep=colon,strut=off} 
\floatstyle{ruled}
\newfloat{listing}{tb}{lst}{}
\floatname{listing}{Listing}
\title{Align to Structure: Aligning Large Language Models \\ with Structural Information}
\author {
    Zae Myung Kim\textsuperscript{\rm 1}\thanks{Work done during internship at Amazon.},
    Anand Ramachandran\textsuperscript{\rm 2},
    Farideh Tavazoee\textsuperscript{\rm 2},\\
    Joo-Kyung Kim\textsuperscript{\rm 2},
    Oleg Rokhlenko\textsuperscript{\rm 2},
    Dongyeop Kang\textsuperscript{\rm 1}
}
\affiliations {
    \textsuperscript{\rm 1}University of Minnesota Twin Cities\\
    \textsuperscript{\rm 2}Amazon\\
    \{kim01756, dongyeop\}@umn.edu, \{anramac, fayt, jookyk, olegro\}@amazon.com
}

\begin{document}

\maketitle

\begin{abstract}
Generating long, coherent text remains a challenge for large language models (LLMs), as they lack hierarchical planning and structured organization in discourse generation. We introduce \textit{Structural Alignment}, a novel method that aligns LLMs with human-like discourse structures to enhance long-form text generation. By integrating linguistically grounded discourse frameworks into reinforcement learning, our approach guides models to produce coherent and well-organized outputs. We employ a dense reward scheme within a Proximal Policy Optimization framework, assigning fine-grained, token-level rewards based on the discourse distinctiveness relative to human writing. 
Two complementary reward models are evaluated: the first improves readability by scoring \textbf{surface-level textual features} to provide explicit structuring, while the second reinforces deeper coherence and rhetorical sophistication by analyzing \textbf{global discourse patterns} through hierarchical discourse motifs, outperforming both standard and RLHF-enhanced models in tasks such as essay generation and long-document summarization.
\end{abstract}

\begin{links}
    \link{Code}{https://github.com/minnesotanlp/StructAlign}
    \link{Datasets}{https://huggingface.co/datasets/zaemyung/writing_prompts_collection}
    \link{Extended version}{https://arxiv.org/abs/2504.03622}
\end{links}

\section{Introduction}\label{sec:intro}
The rapid advances in large language models (LLMs) have sparked significant interest in aligning these models with desired behaviors and outputs \citep{casper2023open,kaufmann2024surveyreinforcementlearninghuman}. While alignment techniques, such as Reinforcement Learning from Human Feedback (RLHF) \citep{Christiano2017deep,ouyang2022training}, have effectively improved the ``helpfulness'' and ``harmlessness'' of generated text, most such work focuses on \textit{what humans prefer}, rather than \textit{how humans structure} an eloquent and coherent discourse. 

Human writers naturally employ established text structures such as problem-solution, cause-effect, or descriptive schemes \citep{meyer1975organization}, to maintain local coherence and logical flow. Theoretical frameworks \citep{mann1987rhetorical} further illuminate how arguments, explanations, and narratives link together to ensure global consistency. Incorporating these structural principles into LLM training could enhance text generation, enabling models to produce more human-like, coherent discourse and better align with human writing conventions.

\begin{figure}[t]
\begin{center}
\includegraphics[width=1.0\columnwidth,trim={3.4mm 0 4.5mm 0},clip]{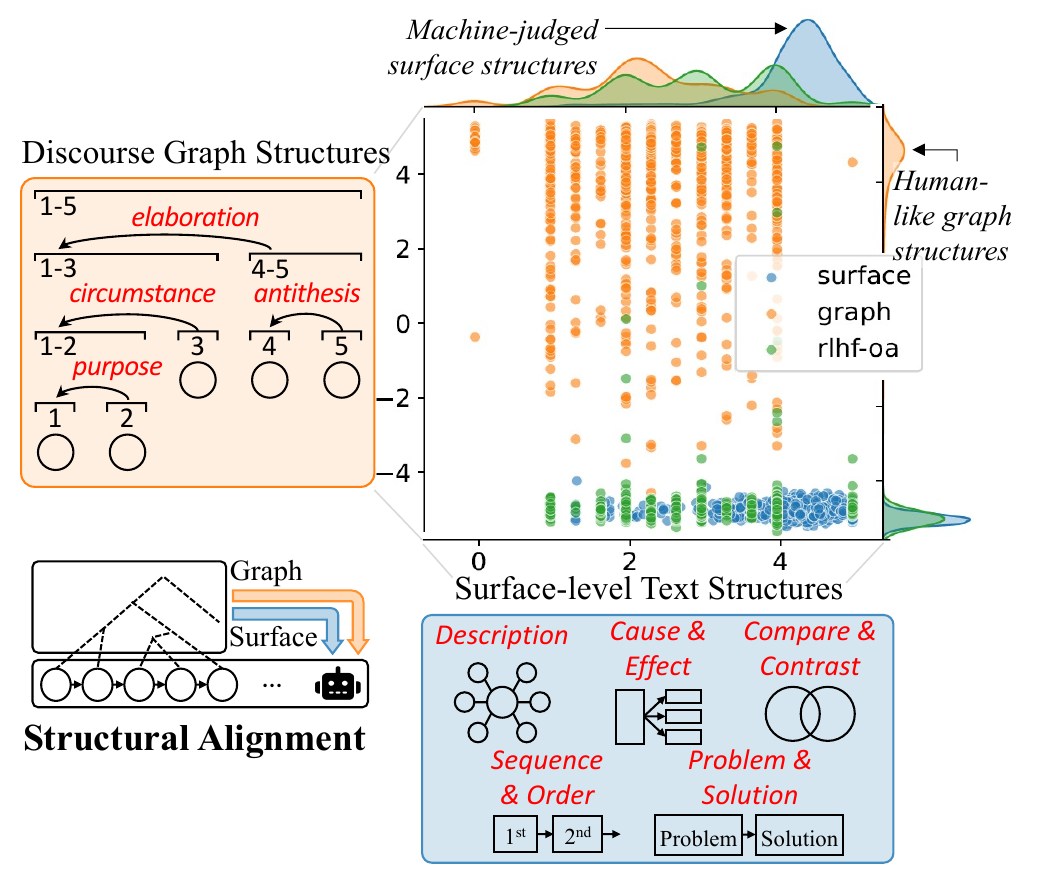}
\end{center}
\caption{Structural alignment extends conventional LLM alignment by incorporating well-defined textual and discourse-level structures, guiding models toward more coherent and human-like text generation.}
\label{fig:struct_align_concept}
\end{figure}

To this end, we introduce \textbf{structural alignment}, a novel alignment objective that extends beyond aligning models with human values to incorporating linguistically-driven document structures.
We explore two avenues of research.
First, we align LLMs with established \textit{surface-level text structures} (cf. Figure \ref{fig:struct_align_concept}) by leveraging an external LLM to guide or evaluate generation within a Proximal Policy Optimization (PPO) framework \citep{schulman2017proximalpolicyoptimizationalgorithms} under a Reinforcement Learning from AI Feedback (RLAIF) paradigm \citep{Bai2022ConstitutionalAH}.

Second, we parse documents using the RST discourse framework to construct \textit{hierarchical discourse trees} and derive discourse motifs--characteristic substructures that highlight patterns distinguishing human-authored from LLM-generated texts.
By integrating an authorship classifier that leverages these motifs to discriminate between human and AI texts, we obtain a (soft) binary label serving as a reward in the PPO framework, effectively guiding the generation process toward more structurally coherent and human-like discourse.

A key motivation for structural alignment lies in \textbf{long-form text generation}, where ensuring logical progression and thematic consistency poses major challenges. Although existing alignment efforts optimize user preferences, they often overlook the ``rhetorical scaffolding'' essential for a comprehensive and logically consistent composition \citep{Mann2006-MANRST}. In contrast, structural alignment focuses on organizing content according to recognized discourse relations, thereby facilitating continuity and depth over extended passages. This emphasis proves particularly beneficial for longer texts---such as multi-paragraph essays or creative narratives---where abrupt topic shifts, undergeneration, or superficial cohesion can profoundly affect the overall reading experience \citep{10.1145/3544548.3581225}. To enhance stability in long-form generation, we introduce a dense reward scheme for PPO that allocates rewards to words based on their relative distinctiveness in human-written and LLM-generated text.
Our contributions are as follows:
\begin{enumerate}
\itemsep0em
    \item Introduce a novel alignment framework by systematically incorporating linguistically grounded structures into the alignment process, enabling LLMs to generate more well-structured, logically consistent text.
    \item Propose a dense reward scheme based on linguistic structures to enhance the stability of RLHF training, particularly for long-form generation (more than 1,000 tokens).
    \item Both quantitative and qualitative evaluations show that structurally aligned LLMs generate discourse structures that more closely mirror human writing, leading to essays with improved clarity, coherence, and overall quality.
\end{enumerate}

\section{Related Work}

\subsection{Aligning Large Language Models}\label{sec:rel_work_alignment}
Various alignment frameworks have emerged recently, such as reinforcement learning from human feedback (RLHF) \citep{Christiano2017deep, ouyang2022training} or from AI Feedback (RLAIF) \citep{Bai2022ConstitutionalAH,ankner2024critiqueoutloudrewardmodels,ye2024improvingrewardmodelssynthetic}, using different optimization techniques such as Direct Preference Optimization (DPO) \citep{dpo}, Group Relative Policy Optimization (GRPO) \citep{shao2024grpo}, among others.
Inspired by RLAIF that leverages LLM judges as reward models (RMs) \citep{ankner2024critiqueoutloudrewardmodels, ye2024improvingrewardmodelssynthetic}, we adopt a similar approach by utilizing an off-the-shelf LLM as the RM, eliminating the need for additional fine-tuning.
Although these alignment techniques differ in their underlying mechanisms, their primary objective remains very similar: shaping model outputs to align with user preferences, social norms, and ethical constraints, ensuring they are perceived as appropriate and acceptable by humans.
In contrast, our work proposes an alternative alignment objective—training LLMs to generate text with well-structured, human-like discourse. Our concept of structural alignment defines an alignment goal rather than prescribing a specific training technique, allowing for flexible implementation across different learning paradigms.

\subsection{Challenges in Long-Form Text Generation}
Recent advances in memory-efficient architectures \citep{hu2021loralowrankadaptationlarge,dao2023flashattention2fasterattentionbetter} and extended context handling \citep{press2022trainshorttestlong,su2023roformerenhancedtransformerrotary} have improved LLMs’ ability to process long-form inputs.
However, producing high-quality long-form text remains challenging \citep{deng-etal-2022-model,10.1145/3544548.3581225}, due to the scarcity of large-scale human-annotated datasets \citep{köksal2024longformeffectiveinstructiontuning}, the inherent subjectivity of extended text evaluation \citep{tan2024proxyqaalternativeframeworkevaluating,wu2025longgenbenchbenchmarkinglongformgeneration}, and the complexities of RL-based alignment \citep{greenberg2021detecting}.

To mitigate the dataset gap, we leverage linguistically grounded discourse structures and LLM-based evaluation, building on the premise that verifying the correctness of long-form text is easier than generating it \citep{zhang2024generative}.
Prior work has addressed RL training instability by employing dense reward shaping \citep{cao-etal-2024-enhancing,chan2024denserewardfreereinforcement}. In a similar vein, we introduce a discourse-aware dense reward scheme (\S\ref{sec:app_reward_shaping}) that assigns rewards based on the structural alignment of generated text with human-like discourse patterns \citep{kim-etal-2024-threads}, precomputed from a large corpus.

\subsection{Structural Organization in Texts}
Text structures---such as problem-solution, cause-effect, comparison-contrast, and description---are foundational in shaping coherent, reader-friendly discourse \citep{meyer1975organization}. They guide the flow of information, highlight relationships between ideas, and ensure logical progression. Meanwhile, hierarchical frameworks like Rhetorical Structure Theory (RST) \citep{mann1987rhetorical} and Segmented Discourse Representation Theory (SDRT) \citep{lascarides2007segmented} add deeper layers of structure by explicitly defining the roles and connections among discourse segments.

\begin{figure*}[th]
\begin{center}
\includegraphics[width=\textwidth]{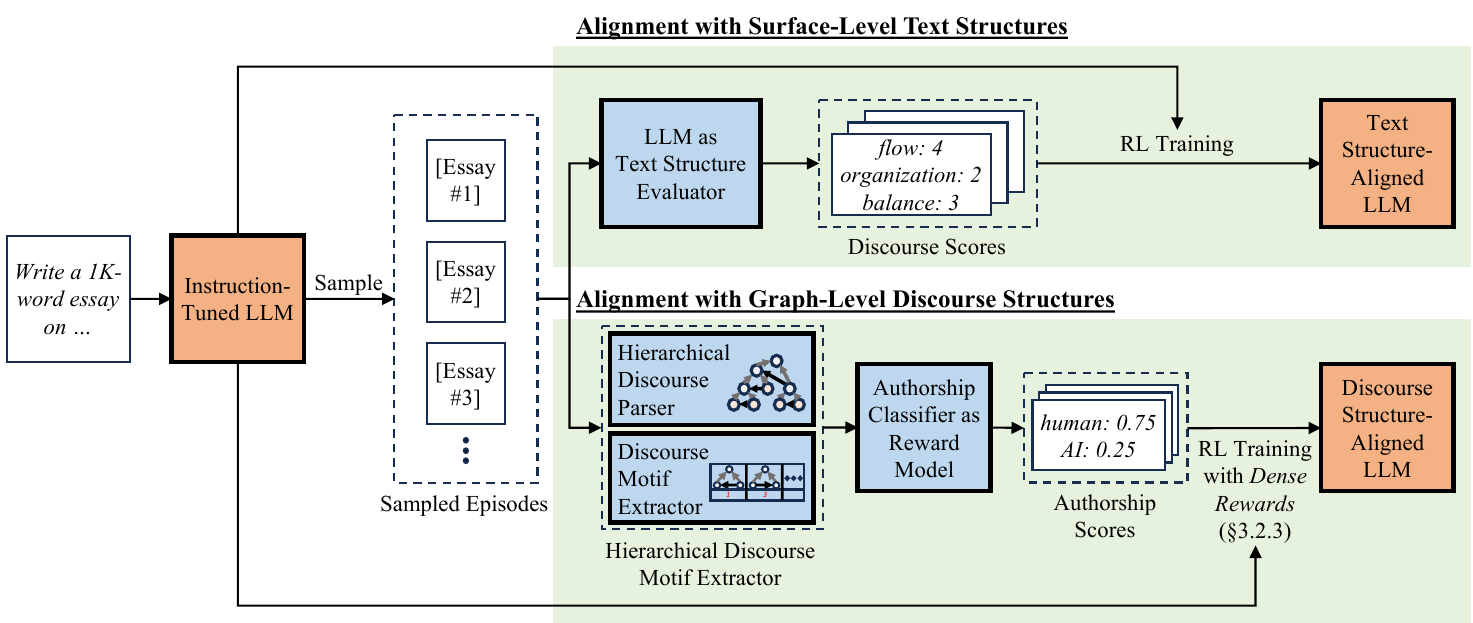}
\end{center}
\caption{An illustration of our approach to structural alignment. We explore two types of reward modeling: (1) scoring surface-level text structures, and (2) scoring human-like discourse via authorship classifier. Components highlighted in blue remain frozen during RL training, while those in orange are trained.}
\label{fig:main_approach}
\end{figure*}

In this work, we adopt RST \citep{mann1987rhetorical} to define a hierarchical discourse tree, wherein Elementary Discourse Units (EDUs) represent basic phrases, and higher-level nodes merge these units into increasingly complex structures. Each connection (edge) carries a discourse label, such as elaboration, and contrast, revealing how ideas fit together. Crucially, these labeled nodes indicate the text's main ideas (nuclei) and supporting details (satellites), illustrating the text's overall rhetorical organization.
Such higher-level structures function as meta-level signals that convey speaker intent, pragmatic cues, and rhetorical strategies \citep{haugh2012speaker,degen2023rational}. By aligning LLMs with these discourse structures, we enable them to maintain thematic continuity, manage contextual shifts, and generate text that is more consistent, context-sensitive, and human-like.

\section{Structural Alignment}\label{sec:approach}
The goal of structural alignment is to enable LLMs generate more logically coherent and human-like text. 
We emphasize that we are introducing a new alignment ``task,'' not a new algorithm, as methods like DPO or GRPO can be readily integrated into our framework.
In this study, we achieve this objective through the RLAIF approach within the PPO framework (\S\ref{sec:rlaif_prelim}, \S\ref{sec:rlaif_struct}), exploring two research avenues: (1) employing an evaluator LLM to score surface-level text structures (\S\ref{sec:app_text_structure}), and (2) evaluating human-like discourse via an authorship classifier grounded in hierarchical discourse motifs (\S\ref{sec:app_graph_structure}). Additionally, we propose a reward-shaping scheme that leverages dense discourse motifs to enhance training stability (\S\ref{sec:app_reward_shaping}).

\subsection{Preliminaries for RLAIF}\label{sec:rlaif_prelim}
Reinforcement Learning from AI Feedback (RLAIF) extends the paradigm of using human feedback by employing AI systems to provide evaluative signals for training a policy model \citep{Bai2022ConstitutionalAH}. Under the Proximal Policy Optimization (PPO) framework \citep{schulman2017proximalpolicyoptimizationalgorithms}, RLAIF optimizes a policy $\pi_\theta(a_t | s_t)$, parameterized by $\theta$, to produce actions $a_t$ (such as generated stories) given states $s_t$ (such as writing prompts). The AI-generated feedback is translated into scalar reward signals $r_t$ that quantify the quality of the actions.

The optimization objective in PPO is designed to maximize expected rewards while ensuring that policy updates are stable and do not deviate excessively from the previous policy. This is achieved by maximizing a clipped surrogate objective function:

\begin{equation}
\begin{aligned}
L^{\text{PPO}}(\theta) = \mathbb{E}_t\left[ \min \left( r_t(\theta) \hat{A}_t, \right. \right. \\
\text{clip}\left. \left. \left( r_t(\theta), 1 - \epsilon, 1 + \epsilon \right) \hat{A}_t \right) \right]
\end{aligned}
\end{equation}

where \(r_t(\theta) = \frac{\pi_\theta(a_t | s_t)}{\pi_{\theta_{\text{old}}}(a_t | s_t)}\) represents the probability ratio between the new and old policies, and \(\epsilon\) is a hyperparameter that limits the extent of policy updates. The advantage function \(\hat{A}_t\) estimates the relative benefit of an action compared to the average, and can be computed as:

\begin{equation}
\hat{A}_t = r_t + \gamma V_{\theta_{\text{old}}}(s_{t+1}) - V_{\theta_{\text{old}}}(s_t)
\end{equation}

with \(V_{\theta_{\text{old}}}(s_t)\) being the value function of the old policy and \(\gamma\) the discount factor. By iteratively performing gradient descent on \(L^{\text{PPO}}(\theta)\), the policy parameters \(\theta\) are updated to maximize the expected rewards:

\begin{equation}
\theta_{\text{new}} = \theta_{\text{old}} + \alpha \nabla_\theta L^{\text{PPO}}(\theta)
\end{equation}

where \(\alpha\) is the learning rate. This iterative process allows the policy to learn from AI feedback, progressively improving its performance while maintaining stability through the clipping mechanism in the PPO objective. Consequently, the policy becomes adept at generating higher-quality outputs as evaluated by AI feedback, reducing the reliance on human annotations and enabling scalable training.

\subsection{RLAIF for Aligning to Structure}\label{sec:rlaif_struct}

Directly applying RLAIF to long-form text generation poses unique challenges, primarily due to the difficulty of \textit{measuring} coherence and organization at scale. Longer outputs require complex thematic progression and consistent rhetorical flow, making it hard to distill quality into a single, easily computed reward. Even powerful off-the-shelf LLMs can struggle to provide reliable ``one-shot'' scores for coherence when dealing with multi-paragraph or multi-page content.

Consequently, we adopt two distinct yet complementary scoring approaches, as illustrated in Figure \ref{fig:main_approach}. The first, introduced in Section \ref{sec:app_text_structure}, uses an evaluator LLM as a Text Structure Evaluator that assigns a 1–5 rating based on surface-level text structures—though an imperfect proxy, it still captures key signals such as logical transitions and topical balance. The second, described in Section \ref{sec:app_graph_structure}, offers a more explicit framework by segmenting extended text and extracting theory-grounded ``discourse motifs'' via RST, which are then fed into an Authorship Classifier for binary classification.

\subsubsection{Surface-Level Text Scoring}\label{sec:app_text_structure}

Our first approach to reward modeling employs a powerful off-the-shelf LLM to provide three surface-level text scores grounded in pragmatics and discourse analysis: Logical Flow and Structure; Hierarchical Organization; and Balance and Emphasis. These scores guide the LLM to generate text that is not only grammatically correct but also pragmatically effective in conveying its intended message within the discourse context.

\begin{itemize}
\itemsep0em
    \item \textit{Logical Flow and Structure} assesses whether ideas progress logically and the overall organization is coherent, ensuring that the discourse is easy to follow.
    \item \textit{Hierarchical Organization} evaluates how effectively content is structured, transitioning from general concepts to specific details, with each section building upon the previous one to support the main argument.
    \item \textit{Balance and Emphasis} examines whether key ideas are appropriately highlighted and whether different points receive balanced coverage, aligning with pragmatic principles of relevance and informativeness.
\end{itemize}
Each score is assigned an integer value from 0 (lowest quality) to 5 (highest quality), and the final reward is computed as their average. The full prompt instruction for this scoring mechanism is provided in supplementary materials.

\subsubsection{Graph-Level Discourse Scoring}\label{sec:app_graph_structure}

\begin{figure}[th]
\begin{center}
\includegraphics[width=0.9\columnwidth]{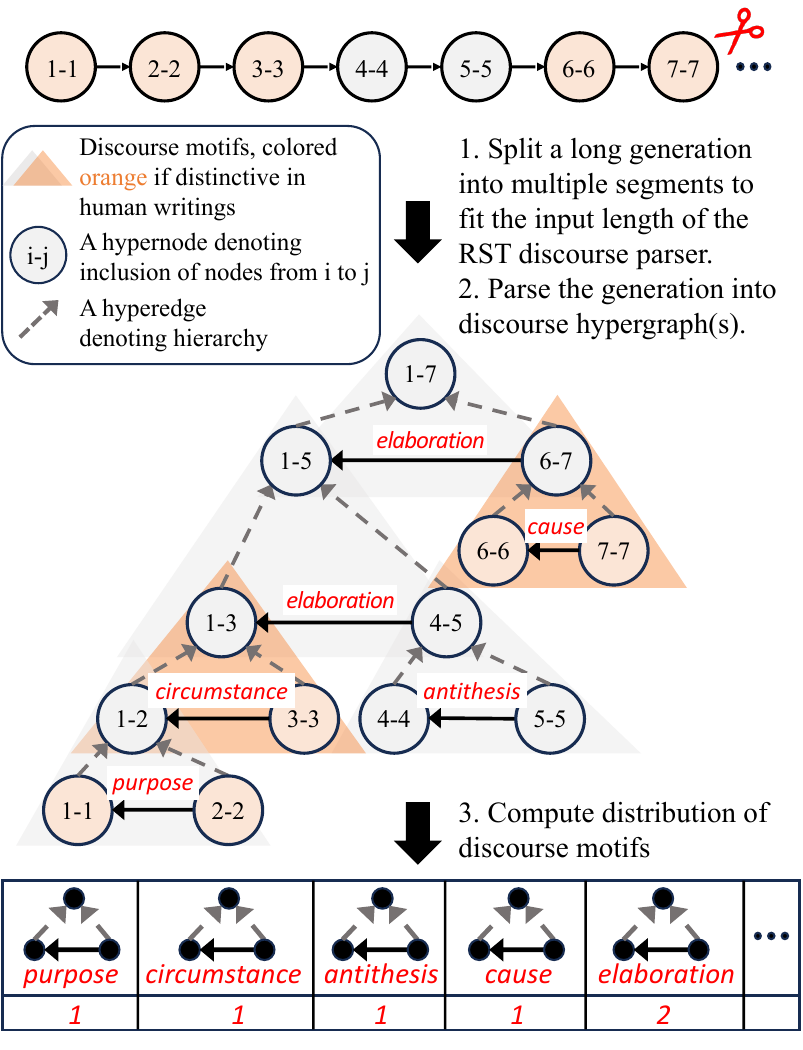}
\end{center}
\caption{The process of extracting discourse motif vectors from generated text, as illustrated in \citet{kim-etal-2024-threads}. We extend the original approach by introducing a segmentation step for texts exceeding the input length of the RST parser. Motif vectors are then aggregated across multiple segments.}
\label{fig:graph_level_align}
\end{figure}

Our approach to aligning LLMs with human-like text structure builds on \citet{kim-etal-2024-threads}, who developed a methodology for distinguishing human-written from machine-generated text using hierarchical discourse analysis. 
Their approach, illustrated in Figure \ref{fig:graph_level_align} employs the RST framework to parse texts into discourse trees, converts these trees into recursive hypergraphs, and extracts discourse motifs---recurring structural patterns---as distinguishing features. These motif distributions serve as inputs for classifiers that effectively detect machine-generated text, even in out-of-distribution and paraphrased cases.

Building on this framework, we extend the RST discourse parser \citep{liu-etal-2021-dmrst} to accommodate longer contexts by segmenting texts at the paragraph level, with each chunk containing approximately 400–512 tokens---a technical limitation that future work can address. Discourse parsing is applied to each segment to compute distributions of distinctive discourse motifs, capturing global discourse patterns beyond surface-level text. To prevent over-representation of motifs, segments are non-overlapping. These aggregated motif distributions are fed into a Longformer-based \citep{beltagy2020longformerlongdocumenttransformer} authorship classifier \citep{li-etal-2024-mage}, where they are concatenated with the model's \textsc{[CLS]} embedding to integrate textual content with explicit structural signals and produce binary classification scores.

Aside from segment-level adaptations, minimal architectural changes are made. The classifier is trained on existing datasets supplemented by curated human-written essays described in Section \ref{sec:dataset}. By capturing how human-authored texts typically display richer and more varied structural patterns, while machine-generated texts often rely on uniform, sequential prediction strategies, this method remains effective even against paraphrasing attacks that obscure surface cues. The classifier's binary output—differentiating human from machine-authored text—becomes the reward signal in our structural alignment task. This setup allows the model to harness hierarchical discourse structures, thereby retaining global coherence patterns and enhancing authorship classification for long-form text generation.

\subsubsection{Reward Shaping with Discourse Motifs}\label{sec:app_reward_shaping}
In the standard RLHF framework with PPO, the episodic reward is assigned only at the end of an episode, specifically, to the final token of the model's generation. This sparsity of rewards limits the training signal's granularity, providing no direct feedback on the quality of individual actions (i.e., token selections) throughout the sequence.

Consequently, as RL episode lengths increase, training stability declines due to the delayed credit assignment, higher exploration complexity, and compounding execution errors \citep{pignatelli2024a}. Reward variance also grows, making policy updates noisier and less effective, particularly in policy gradient methods. While techniques like curiosity-driven exploration \citep{pathak2017curiositydrivenexplorationselfsupervisedprediction}, imitation learning \citep{reddy2019sqilimitationlearningreinforcement}, or curriculum learning \citep{NEURIPS2021_503e7dbb} can help, we instead employ reward shaping to provide more fine-grained signals—taking advantage of the inherently granular structure of language, in particular, discourse trees.

\begin{figure}[t!]
\begin{center}
\includegraphics[width=\columnwidth]{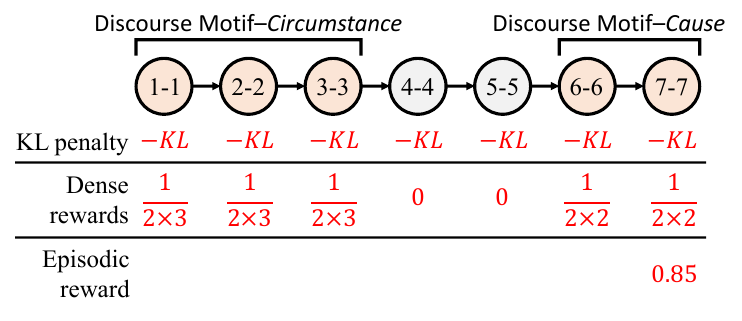}
\end{center}
\caption{In addition to the episodic scalar reward, tokens contributing to human-distinctive discourse motifs receive a reward of $\frac{1}{2 \cdot \text{num\_tokens}}$.}
\label{fig:dense_rewards}
\end{figure}

We identify specific tokens within each sequence that should receive non-zero token-level rewards, enabling more fine-grained guidance for the policy model.
Specifically, in addition to this episodic scalar reward, we assign $\frac{1}{2 \cdot \text{num\_tokens}}$ reward scores to tokens that contribute to a discourse motif that is more distinctively characteristic of human writing compared to those frequently observed in LLM-generated text (cf. Figure \ref{fig:dense_rewards}). These distinctive discourse motifs are identified across entire corpora by computing Motif Frequency-Inverse Document Frequency (MF-IDF) scores and selecting motifs that exceed a threshold of at least one standard deviation, following the approach of \citet{kim-etal-2024-threads}. To preserve finer granularity, we compute discourse motifs over EDUs rather than full sentences; each EDU contains, on average, 20 subword tokens.

While the overall PPO training objective remains unchanged, the implementation involves adding these token-level rewards to the corresponding positions in the reward tensor for each sequence. This design ensures gradient updates capture the localized contribution of each token, improving credit assignment throughout generation. As a result, policy improvements typically occur more quickly and with greater stability than when relying on a single end-of-sequence reward \citep{chan2024denserewardfreereinforcement}.

\subsection{Length-Penalty Normalization of Rewards}\label{sec:length_pen}

Noting that existing LLMs often fail to adhere to the desired response length specified in prompts \citep{wu2025longgenbenchbenchmarkinglongformgeneration}, we adjust the original score by applying a penalty proportional to the shortfall in response length, but only if the response is shorter than the desired length.

The normalized score, $S_{\text{n}}$, based on the response length can be represented as:

\begin{equation}    
S_{\text{n}} = S_{\text{o}} \times \left[1 - \alpha \times \max\left(0, \dfrac{L_{\text{d}} - L_{\text{r}}}{L_{\text{d}}}\right)\right]
\end{equation}

where $S_{\text{o}}$ is the original score; $L_{\text{r}}$ is the length of the generated response; $L_{\text{d}}$ is the desired length of the response; and $\alpha$ is the length penalty factor.

\section{Experiments}

As shown in Figure \ref{fig:main_approach}, we investigate two reward modeling approaches within the RLAIF framework using PPO: (1) surface-level text structure scores and (2) graph-level discourse structure scores. To support this, we construct a dataset of prompts by collecting and refining writing prompts for essay generation (\S\ref{sec:dataset}) and validate the quality of the LLM-based Text Structure Evaluator (\S\ref{sec:verify_rm}). Finally, we evaluate structure-aligned LLMs against baseline models trained with standard RLHF using an off-the-shelf general-purpose RM (\S\ref{sec:struct_align_exp}).

Throughout our experiments, we use \textsc{Qwen2-72B-Instruct-AWQ} as the Text Structure Evaluator and \textsc{Qwen2-1.5B-Instruct} as the initial policy model for structural alignment \citep{yang2024qwen2technicalreport}. 
Due to resource constraints, we selected a 1.5B parameter model as it strikes a balance between being sufficiently powerful and small enough to enable efficient PPO training in data parallelism mode. At the time of our experiments, Qwen 2 models offered state-of-the-art performance in the LLM landscape.
For hierarchical discourse analysis, we adopt the DMRST parser \citep{liu-etal-2021-dmrst} as our Hierarchical Discourse Parser, while the Discourse Motif Extractor and Authorship Classifier are based on \citet{kim-etal-2024-threads}. Details on implementation are provided in Appendix \ref{app:exp_details}.

\subsection{Constructing an Essay Prompt Dataset}\label{sec:dataset}
Our work focuses on formal essay writing, as its well-defined rhetorical conventions provide a strong foundation for evaluating structural alignment. To ensure diversity in formal essays, we collected prompts from various domains emphasizing clear logical structures and rhetorical coherence.

Specifically, our PPO training dataset is constructed by refining and integrating multiple existing datasets containing essay prompts, including those from an English proficiency exam, persuasion corpus \citep{crossley2023large}, and an argumentation source, namely, the Change My View (CMV) subreddit \citep{10.1145/2872427.2883081}.

For CMV-derived samples, we employ an LLM\footnote{Qwen/Qwen2-72B-Instruct} to generate essay instructions that require students (or LLMs) to either support or oppose the original poster's opinion. Each CMV discussion yields two types of instructions: one prompting the student to argue in favor of the given claim and the other requiring them to argue against it. Given the nature of CMV discussions, many of these prompts may be inherently controversial.

To ensure an effective evaluation of model performance, 30\% of the dataset is randomly sampled as the test set, except for CMV, where a predefined test split is already provided. The final dataset consists of 26,013 samples for training and 4,096 samples for testing.\footnote{Included in the supplementary material.}




\subsection{Verification of Text Structure Evaluator}\label{sec:verify_rm}
To validate the quality of the text scores generated by our LLM-based RM, Text Structure Evaluator, we examined their correlation with human scores using the ``Hewlett Foundation: Automated Essay Scoring'' dataset \citep{asap-aes}.\footnote{We exclude verification of the graph-level discourse scoring approach, as it follows the validated method proposed by \citet{kim-etal-2024-threads}.} This dataset contains high school student essays along with scores assigned by expert human graders. For each essay, students responded to a writing prompt, and human annotators assessed their quality.

When analyzing the Pearson correlation between the discourse scores from the surface-level RM and human-assigned scores, we observe moderate positive correlations across key dimensions: coherence (0.47), organization (0.44), balance (0.39), purpose (0.37), and variability (0.39). These values indicate the discourse scores align reasonably well with human judgments in these areas. Additionally, using linear regression, we measured the mean squared error of the predicted scores to be 0.79 within the 0-4 range, translating to an average deviation of approximately 0.88 points from human scores. We determined the correlation results were strong enough to justify using the LLM-based discourse scoring as the RM and proceeded with PPO training.

\subsection{Structural Alignment}\label{sec:struct_align_exp}
We evaluate the effectiveness of structural alignment by assessing its impact on the quality of generated essays and the similarity of generated summaries to human-written long summaries.

We use ``Base'' to denote the base model, \textsc{Qwen2-1.5B-Instruct}. Models $\text{SA}_{\text{S}}$ and $\text{SA}_{\text{G}}$ represent structurally aligned variants of the base model, where alignment is assessed using surface-level textual scores and graph-level discourse structures, respectively. Lastly, $\text{RLHF}_{\text{OA}}$ denotes the base model aligned with an off-the-shelf RM from OpenAssistant\footnote{OpenAssistant/reward-model-deberta-v3-large-v2}, which was trained to predict human-preferred responses given a question.

\subsubsection{Exp \#1: Assessment of Generated Essay Quality}

\paragraph{Setup.}
We configure the models to generate sequences between 400 and 2K tokens. As training progresses within each epoch, the required generation length is gradually increased using a length penalty reward shaping mechanism (\S\ref{sec:length_pen}). For this experiment, we introduce two additional models by further training $\text{SA}_{\text{S}}$ and $\text{SA}_{\text{G}}$ using the alternate method in a two-step process, resulting in $\text{SA}_{\text{S}\rightarrow\text{G}
}$ and $\text{SA}_{\text{G}\rightarrow\text{S}}$.
We evaluated generation quality by randomly sampling 1,000 writing prompts from the test set and generating 700-token essays with our models. Because no gold-standard reference essays exist for these prompts, reference-based evaluation is not feasible. Additionally, assessing such lengthy text manually can be both time-consuming and prone to inconsistencies at longer contexts. In response, we adopted an LLM-as-judge approach, which is increasingly common in the field \citep{chiang-lee-2023-large}. Specifically, we employed \textsc{GPT-4o-2024-08-06} to assess overall essay quality, presenting all candidate generations in a randomized order to reduce potential bias in ordering.

\begin{figure}[t]
\begin{center}
\includegraphics[width=0.85\columnwidth]{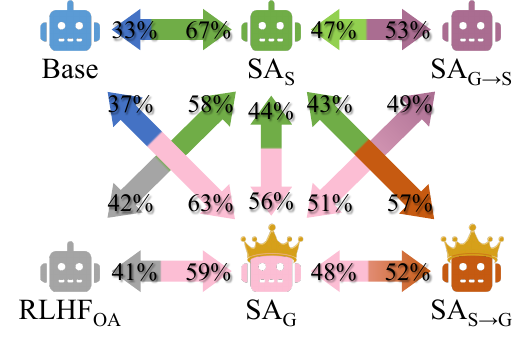}
\end{center}
\caption{Pairwise evaluation results for six models.
$ \text{Base} $ is the unaligned Qwen2-1.5B-Instruct model.
$ \text{SA}_{\mathrm{S}} $ and $ \text{SA}_{\mathrm{G}} $ are Base aligned with surface-level and graph-level structural rewards, respectively.
$ \text{RLHF}_{\mathrm{OA}} $ is Base aligned with an OpenAssistant RLHF reward model.
$ \text{SA}_{\mathrm{S} \rightarrow \mathrm{G}} $ and $ \text{SA}_{\mathrm{G} \rightarrow \mathrm{S}} $ are the two-stage variants, applying the opposite structural reward in a second alignment step.}
\label{fig:eval_llm_judge_base}
\end{figure}

\paragraph{Results.}
Figure \ref{fig:eval_llm_judge_base} presents evaluation results comparing generation quality across the various alignment settings.
Overall, results show that models trained with structural alignment consistently outperform their baselines, with graph-level alignment providing the most substantial improvements. The two-step alignment process yields slightly higher performance; however, the improvement appears marginal relative to the additional computational cost. Since the two structural alignment methods exhibit low correlation (\S\ref{para:corr_s_g}), we believe their effects are complementary. Developing more effective multi-reward alignment strategies could further improve performance, making this a promising avenue for future research.
We provide two samples of generated essays with target lengths of 350 and 1K tokens in supplementary materials. Overall, the learned policy models effectively utilize explicit discourse connectives and frequently attempt to structure the text by dividing it into sections.


\subsubsection{Exp \#2: Long-Document Summarization}
\paragraph{Setup.} We evaluate the aligned models on the downstream task of summarizing a long document. Specifically, we utilize the \textsc{GovReport} dataset \cite{huang-etal-2021-efficient} which is a large-scale dataset comprising around 19.5K US government reports, each paired with expert-written abstractive summaries.
The dataset contains significantly longer documents (averaging 9.4K words) and summaries (553 words) compared to datasets like PubMed and arXiv.
We randomly selected 5K reports from the U.S. Government Accountability Office (GAO), ensuring that each report selected for summarization does not exceed 14K tokens in length.
To guide the LLM via prompting, we identify key aspects to summarize by analyzing the section titles of the gold summaries (``Highlights''), which typically focus on two main elements: \textbf{why} the report was conducted and \textbf{what} it found.

\begin{table}[t!]
\centering\small
\resizebox{0.67\columnwidth}{!}{%
\begin{tabular}{l|cccc}
\toprule
Model                                       & R-1   & R-2   & R-L   \\
\midrule
Base                                        & 53.21 & 20.13 & 50.39 \\
Base+$\text{RLHF}_{\text{OA}}$              & 53.25 & 20.25 & 50.47 \\
Base+$\text{SA}_{\text{S}}$           & 55.45 & 21.43 & 52.30 \\
Base+$\text{SA}_{\text{G}}$             & 55.86 & 21.72 & 52.81 \\
\bottomrule
\end{tabular}%
}
\caption{Evaluations on long document summarization (\textsc{GovReport}). R-1 and R-2 measure unigram and bigram overlap, respectively. R-L leverages the longest common subsequence to assess sentence-level structural similarity.}

\label{tab:eval_summ}
\end{table}

\paragraph{Results.}
Table \ref{tab:eval_summ} presents the ROUGE scores \citep{lin-2004-rouge} for the generated summaries, demonstrating that structural alignment significantly enhances long-document summarization. Among the approaches, graph-based alignment once again delivers the best overall performance.

\subsection{Analyses}\label{sec:analyses}
By incorporating structured reward signals, we observe significant improvements in discourse organization and human-like text structuring.

\paragraph{Effects of Surface-Level Structural Alignment} 

\begin{figure}[t!]
\begin{center}
\includegraphics[width=1.0\columnwidth,trim={2mm 0 4mm 0},clip]{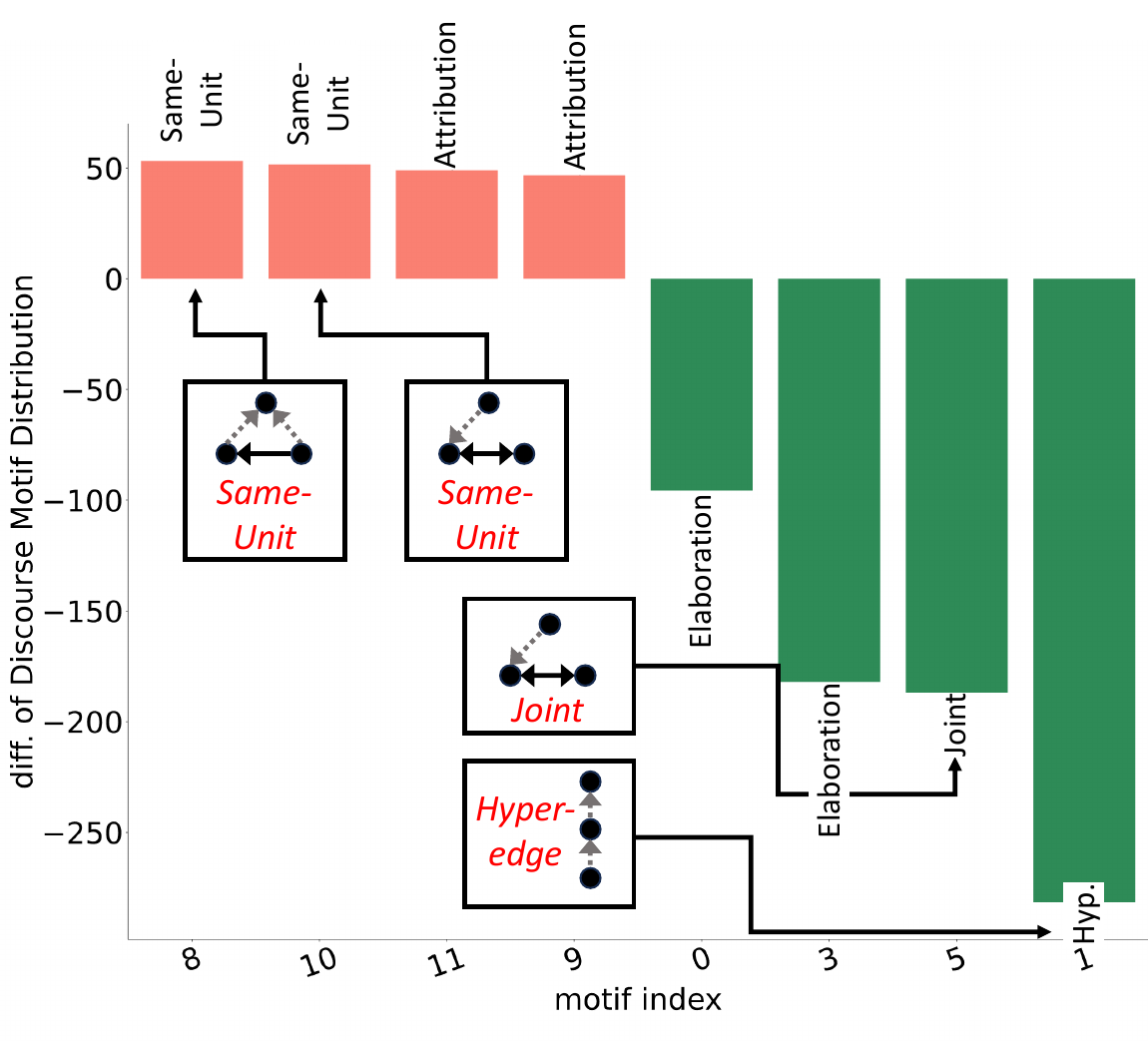}
\end{center}
\caption{Difference in discourse motif distributions before (red) and after (green) surface-level text structure alignment. Notably, we observe a significant increase in the usage of hierarchical relations such as ``Joint'' and ``Hyperedges (Hyp.)''. Full version is in supplementary materials.}
\label{fig:before_and_after}
\end{figure}

We qualitatively observe that aligning LLMs with surface-level text structure scores results in (1) increased use of discourse connectives (e.g., \textit{therefore}, \textit{however}, \textit{in contrast}, and \textit{moreover}), (2) improved logical progression and argumentative flow, and (3) more frequent inclusion of section headings (e.g., \textit{Introduction}, \textit{Claim 1}, \textit{Claim 2}, and \textit{Conclusion}). A similar pattern is indirectly observed in Figure \ref{fig:before_and_after}, where surface-level text structure alignment leads to a significant increase in discourse motifs that exemplify hierarchical structures, such as ``Joint'' and ``Hyperedges.''

\paragraph{Effects of Graph-Level Structural Alignment} 

\begin{figure}[t!]
\begin{center}
\includegraphics[width=0.93\columnwidth,trim={3mm 0 17mm 0},clip]{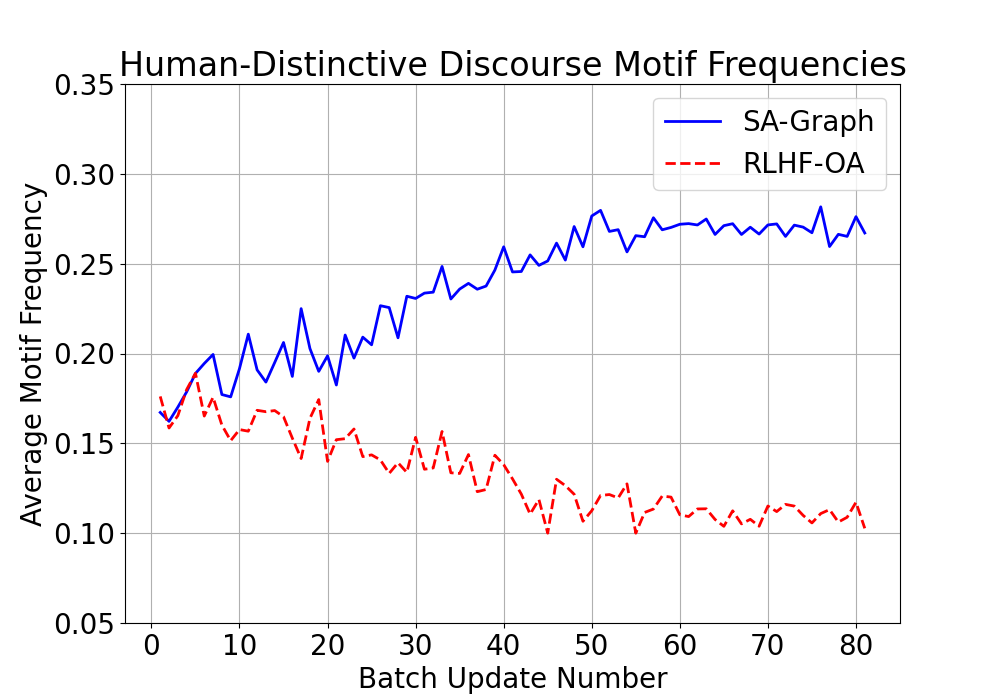}
\end{center}
\caption{Comparison of human-distinctive discourse motif frequencies across batch updates. The blue line (our method) shows an increasing trend, while the red dashed line (standard PPO) fluctuates downward.}
\label{fig:comp_avg_motif_freq}
\end{figure}

Unlike surface-level alignment, graph-level discourse structure alignment focuses on \textbf{global discourse organization} rather than local discourse markers (e.g., connectives). Figure \ref{fig:comp_avg_motif_freq} tracks the proportion of human-distinctive discourse motifs identified in each training batch, illustrating how these motifs evolve throughout the generation process. The plot shows that (i) their proportion steadily increases during training, and (ii) the trend plateaus after approximately 50 batch steps, suggesting discourse structuring reaches an optimal threshold.
Notably, standard RLHF training without structural alignment shows a slow decline in such motif proportions, suggesting that conventional RMs do not reinforce human-like discourse organization as effectively.

\paragraph{Correlation between Surface and Graph Scores}\label{para:corr_s_g} 
Figure \ref{fig:struct_align_concept} visualizes the relationship between surface and graph-level logits for essays generated by the two structurally aligned models and the standard RLHF-aligned model. While the structurally aligned models tend to achieve higher scores within their respective evaluation criteria, there appears to be little to no correlation between surface-level and graph-level scores. Additionally, we observe that essays generated by the standard RLHF-aligned model exhibit wide variance in surface scores and are predominantly classified as machine-generated, as indicated by negative graph logits.

In summary, our findings suggest that (1) Surface-level alignment enhances readability and clarity via explicit structuring techniques; (2) Graph-level alignment fosters deeper coherence and rhetorical sophistication by reinforcing global discourse structure; and (3) These two levels of alignment appear to be complementary, with the two-step process improving overall quality.

\section{Conclusion}
We present an RL approach that aligns LLMs with human-like discourse by utilizing two RMs---one for surface fluency, another for global discourse structure---resulting in clearer, more coherent long-form text that outperforms RLHF baselines. Future work includes scaling to larger models and diverse datasets for greater robustness, and incorporating other linguistic frameworks to refine reward signals and enhance performance.

\appendix
\onecolumn

\section{Experiment Details}\label{app:exp_details}
\paragraph{RL Pipeline Setup}
Our Text Structure Evaluator is a 72B-parameter language model, which made an efficient serving framework essential. We used SGLang \citep{zheng2024sglangefficientexecutionstructured} to load the \textsc{Qwen2-72B-Instruct-AWQ} model, where AWQ stands for Activation-aware Weight Quantization \citep{lin2024awqactivationawareweightquantization}. This quantized model can be loaded onto a single NVIDIA A100 80GB GPU. We ran eight separate evaluator instances, each on its own A100 GPU, and communicated with these evaluators through HTTP requests. SGLang handled these batch requests efficiently, allowing us to process multiple queries in parallel.

For training, we used the TRL library \citep{vonwerra2022trl} to perform Proximal Policy Optimization (PPO). Although TRL provides a strong base for reinforcement learning from human or AI feedback, we made extensive modifications to support our online reward mechanism, which involved querying the Text Structure Evaluator in real-time. The modified code, which enables discourse alignment via remote evaluation, will be made publicly available for further research and development.

The actual PPO-based reinforcement learning (RL) training took place on another server with eight NVIDIA A100 80GB GPUs. We set the per-device train batch size to 2, gradient accumulation steps to 4, local rollout forward batch size to 12, and KL coefficient to 0.03. Despite having large VRAM, we could not increase these batch sizes further due to the large output generation lengths (up to 2K tokens) which places substantial memory demands on the system.

\paragraph{Graph-Level Discourse Structure Scoring}
We closely followed the approach outlined by \citet{kim-etal-2024-threads} to implement our Graph-Level Discourse Structure Scoring pipeline. Using their curated list of distinctive discourse motifs, we constructed a Discourse Motif Extractor and trained a Longformer-based \citep{beltagy2020longformerlongdocumenttransformer} authorship classifier as described in the original paper. However, we extended the training set for the classification by adding 5K essay generations from \textsc{Qwen2-1.5B-Instruct} as negative samples, in addition to the datasets originally used in \citet{kim-etal-2024-threads}.

\section{Training Plots}

\begin{figure}[th]
\begin{center}
\includegraphics[width=0.9\columnwidth]{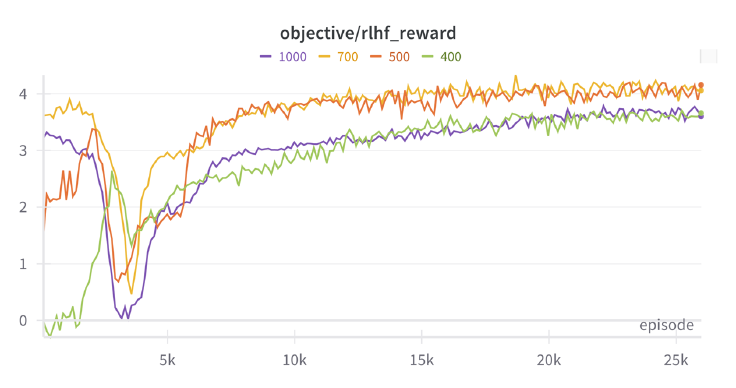}
\end{center}
\caption{The curves represent the average reward scores provided by the RM over multiple steps or episodes, with each curve corresponding to models targeting different output lengths.}
\label{fig:surface_reward_curves}
\end{figure}

Figure \ref{fig:surface_reward_curves} shows the mean RLHF reward for training via the surface-level text structure scoring. The mean reward is the average feedback score over multiple episodes, reflecting how well the model aligns with desired outcomes. We note that different curves represent separate training of the same baseline models with different target lengths of generation. We can see that the policy models learned to optimize its generation after around 7K episodes.

\newpage

\section{Surface-Level Discourse Motifs Pre- and Post Alignment}\label{app:diff_dist_full}
\begin{figure}[th]
\begin{center}
\includegraphics[width=0.95\columnwidth]{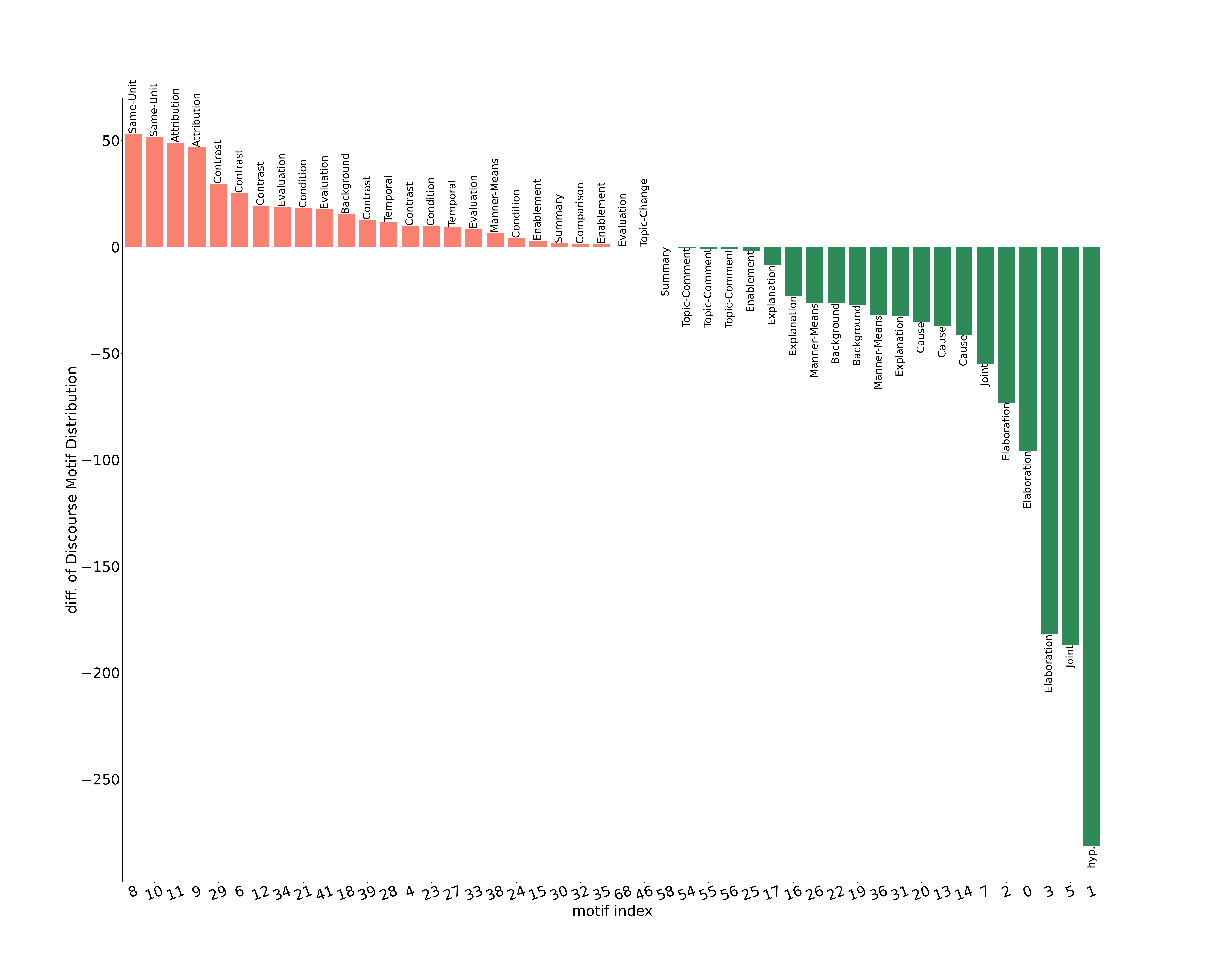}
\end{center}
\caption{Difference of distribution of size-3 discourse motifs.}
\label{fig:discourse_dist}
\end{figure}

\section{Prompt Instruction for LLM as Text Structure Evaluator}
\label{app:prompt_aspect_eval}
\begin{lstlisting}[backgroundcolor=\color{lightgray}]
You will act as an English instructor and evaluate the quality of an essay or story written by a student in response to given instructions. When grading, consider the following discourse aspects of the text.
- Logical Flow and Structure (flow): Assess the logical progression of ideas and the overall organization of the text, ensuring that it is easy to follow and well-structured.
- Hierarchical Organization (organization): Examine the organization of ideas in a hierarchical manner, from general to specific, ensuring that each section supports the main argument or narrative.
- Balance and Emphasis (balance): Ensure that important ideas are appropriately emphasized and that there is a balance in the coverage of different points or sections of the text.

For each aspect, you need to assign an integer score from 0 (worst quality) to 5 (best quality).
When assigning the score, carefully consider which specific parts of the text relate to each aspect.

Assign lower scores when:
- The text is poorly structured and do not conform to the standard of an English essay or a story.
- The text contains a lot of non-sensical words such as special tokens or programming code.
- The text contains a lot of non-English words.
- The text does not fully answer the writing instruction with full content, and therefore, is unfinished.

Your evaluation output should conform to the following JSON format:
{
  "flow": int,
  "organization": int,
  "balance": int
}

Write <EOE> after outputting the JSON result.
\end{lstlisting}

\section{Prompt Instructions for LLM as Discourse Annotator}
\label{app:prompt_discourse_annot}

\begin{lstlisting}[language=python,backgroundcolor=\color{lightgray}]
intro_prompt = """You will act as an English instructor and analyze how a specific {target_type} is crafted to serve a discourse-related purpose in the overall text.
The specific {target_type} is:
```
{target}
```

It is used and can be found in the following scene:
```
{scene}
```
"""

summary_prompt = """First, summarize the {target_type} in one sentence using no more than 70 words. When finished, append <EOE> at the end.
"""

purpose_prompt = """What is the discourse-related purpose of the {target_type}? Answer as a concise sentence using no more than 50 words without repeating the {target_type}. When finished, append <EOE> at the end.
"""

discourse_label_prompt = """Based on the purpose you found, determine which one of the following discourse labels is most applicable. When finished, append <EOE> at the end.
- Temporal: Describes a relationship where events or propositions are sequenced in time, where one follows another chronologically.
- Causation: Refers to a cause-effect relationship where one proposition leads to or results in the other.
- Conditional: Represents a relationship where one proposition is contingent on the fulfillment of another.
- Purpose: One proposition explains the intention or goal behind another.
- Concession: A relationship where one proposition presents an opposing viewpoint or circumstance, but the other still holds.
- Contrast: Highlights a difference between two propositions.
- Similarity: Describes a relationship where two propositions highlight similar characteristics or situations.
- Conjunction: Involves the simple addition of information, where two or more propositions are connected logically.
- Disjunction: Represents alternatives between propositions, where one excludes the other.
Selected label:
"""
\end{lstlisting}

\newpage

\section{Generated Samples for Different Target Lengths}
\label{app:gen_samples}

\subsection{A Writing Prompt Example}
\begin{lstlisting}[backgroundcolor=\color{lightgray}]
Write an essay that argues against the claim that there is no universal right or wrong, and instead, there are just perspectives. In your essay, consider discussing the existence of objective moral truths and the role of reason in discerning right from wrong. Be sure to provide evidence and examples to support your argument.
\end{lstlisting}

\subsection{Target Length of 350 Tokens}
\begin{lstlisting}[backgroundcolor=\color{lightgray}]
The notion that there is no universal right or wrong perspective is a misguided belief that ignores the profound impact of reason and objective moral truths on human behavior and societal development. Historical and contemporary evidence consistently demonstrate the existence of objective moral truths that guide societies towards progress and harmony.

For instance, the Universal Declaration of Human Rights, adopted by the United Nations in 1948, articulates a set of fundamental rights that transcend cultural, economic, and political differences. This document serves as a beacon of hope, guiding nations towards equitable treatment and respect for human dignity. Numerous examples from around the world illustrate how adherence to these principles has led to significant improvements in human rights, economic development, and social cohesion.

Moreover, the role of reason in discerning right from wrong is evident in the scientific method, which has revolutionized fields such as medicine, engineering, and environmental conservation. Empirical evidence and logical reasoning have led to groundbreaking discoveries that have improved human health and quality of life.

However, it is crucial to acknowledge that while objective moral truths exist, they are not absolute. Cultural, historical, and individual perspectives can influence moral judgments and practices. For instance, the widespread acceptance of slavery in the Western world during the 19th century was largely due to prevailing cultural and economic factors rather than objective moral truths.

In conclusion, while there may be no universal right or wrong perspective, objective moral truths continue to guide societies towards progress and human flourishing. Evidence from history and contemporary examples demonstrates the transformative power of reason and collective action in promoting ethical behavior and social harmony.
\end{lstlisting}

\subsection{Target Length of 1K Tokens}
\begin{lstlisting}[backgroundcolor=\color{lightgray}]
The claim that there is no universal right or wrong is a common one, often expressed by those who believe that moral judgments are subjective and vary from person to person. This belief is often referred to as relativism, and it has been a topic of debate for centuries. While some argue that there are objective moral truths that are universally applicable, others contend that morality is a matter of personal opinion and varies from culture to culture.

One of the most prominent proponents of relativism is philosopher John Stuart Mill, who argued that moral judgments are based on individual preferences and that there is no objective standard for what is right or wrong. Mill believed that morality is a matter of personal choice and that individuals should be free to make their own judgments about what is right and wrong based on their own beliefs and values.

However, this view of morality has been criticized for its lack of clarity and for its inability to provide a consistent framework for ethical decision-making. Many argue that morality is not simply a matter of personal preference, but rather it is based on objective moral truths that are shared by all humans.

One example of this is the concept of universal moral principles, such as the prohibition of murder, theft, and lying. These principles are believed to be universally applicable and are considered to be objective moral truths. For example, the prohibition of murder is considered to be a universal moral principle because it is a fundamental aspect of human nature and is recognized by all cultures and societies.

Another example of objective moral truths is the concept of justice. Justice is considered to be a universal moral principle because it is a fundamental aspect of human society and is recognized by all cultures and societies. The idea of justice is based on the belief that everyone should be treated equally and that fairness and equality are essential components of a just society.

Despite the existence of objective moral truths, some argue that morality is still a matter of personal opinion and that there is no objective standard for what is right or wrong. This view is often referred to as moral relativism, and it is based on the idea that moral judgments are based on individual preferences and cultural differences.
One of the main criticisms of moral relativism is that it fails to provide a consistent framework for ethical decision-making. Many argue that moral relativism leads to a lack of consistency and inconsistency in moral judgments, as different cultures and societies may have different moral standards and values.

Another criticism of moral relativism is that it undermines the importance of moral education and the development of moral reasoning skills. Without a clear understanding of objective moral truths, individuals may struggle to make ethical decisions and may be more likely to make mistakes or to act in ways that are harmful to others.

In conclusion, the claim that there is no universal right or wrong is a common one, and it is often expressed by those who believe that moral judgments are subjective and vary from person to person.

\end{lstlisting}
\twocolumn
\section*{Acknowledgments}
We are grateful to the researchers at Amazon and members of the intern cohort for their valuable feedback during weekly research discussions. We also thank members of Minnesota NLP for their insightful input during group meetings. ZMK is generously supported by the 3M Science and Technology Fellowship and the Doctoral Dissertation Fellowship at the University of Minnesota.

\bibliography{aaai2026,anthology}

@inproceedings{Christiano2017deep,
 author = {Christiano, Paul F and Leike, Jan and Brown, Tom and Martic, Miljan and Legg, Shane and Amodei, Dario},
 booktitle = {Advances in Neural Information Processing Systems},
 editor = {I. Guyon and U. Von Luxburg and S. Bengio and H. Wallach and R. Fergus and S. Vishwanathan and R. Garnett},
 pages = {},
 publisher = {Curran Associates, Inc.},
 title = {Deep Reinforcement Learning from Human Preferences},
 url = {https://proceedings.neurips.cc/paper_files/paper/2017/file/d5e2c0adad503c91f91df240d0cd4e49-Paper.pdf},
 volume = {30},
 year = {2017}
}

@incollection{lascarides2007segmented,
  title={Segmented discourse representation theory: Dynamic semantics with discourse structure},
  author={Lascarides, Alex and Asher, Nicholas},
  booktitle={Computing meaning},
  pages={87--124},
  year={2007},
  publisher={Springer}
}

@misc{shao2024grpo,
      title={DeepSeekMath: Pushing the Limits of Mathematical Reasoning in Open Language Models}, 
      author={Zhihong Shao and Peiyi Wang and Qihao Zhu and Runxin Xu and Junxiao Song and Xiao Bi and Haowei Zhang and Mingchuan Zhang and Y. K. Li and Y. Wu and Daya Guo},
      year={2024},
      eprint={2402.03300},
      archivePrefix={arXiv},
      primaryClass={cs.CL},
      url={https://arxiv.org/abs/2402.03300}, 
}

@book{meyer1975organization,
  title={The Organization of Prose and Its Effects on Memory},
  author={Meyer, B.J.F.},
  isbn={9780720493009},
  lccn={75022077},
  series={North-Holland studies in theoretical poetics},
  url={https://books.google.com/books?id=3e9YAAAAMAAJ},
  year={1975},
  publisher={North-Holland Publishing Company}
}

@article{su2023roformerenhancedtransformerrotary,
title={RoFormer: Enhanced Transformer with Rotary Position Embedding}, 
author={Jianlin Su and Yu Lu and Shengfeng Pan and Ahmed Murtadha and Bo Wen and Yunfeng Liu},
year={2024},
volume={568},
journal={Neurocomputing}
}

@inproceedings{10.1145/3544548.3581225,
author = {Mirowski, Piotr and Mathewson, Kory W. and Pittman, Jaylen and Evans, Richard},
title = {Co-Writing Screenplays and Theatre Scripts with Language Models: Evaluation by Industry Professionals},
year = {2023},
isbn = {9781450394215},
publisher = {Association for Computing Machinery},
address = {New York, NY, USA},
url = {https://doi.org/10.1145/3544548.3581225},
doi = {10.1145/3544548.3581225},
booktitle = {Proceedings of the 2023 CHI Conference on Human Factors in Computing Systems},
articleno = {355},
numpages = {34},
location = {Hamburg, Germany},
series = {CHI '23}
}

@article{Mann2006-MANRST,
	author = {William C. Mann and Maite Taboada},
	doi = {10.1177/1461445606061881},
	journal = {Discourse Studies},
	number = {3},
	pages = {423--459},
	publisher = {Sage Publications},
	title = {Rhetorical Structure Theory: Looking Back and Moving Ahead},
	volume = {8},
	year = {2006}
}

@article{casper2023open,
title={Open Problems and Fundamental Limitations of Reinforcement Learning from Human Feedback},
author={Stephen Casper and Xander Davies and Claudia Shi and Thomas Krendl Gilbert and J{\'e}r{\'e}my Scheurer and Javier Rando and Rachel Freedman and Tomek Korbak and David Lindner and Pedro Freire and Tony Tong Wang and Samuel Marks and Charbel-Raphael Segerie and Micah Carroll and Andi Peng and Phillip J.K. Christoffersen and Mehul Damani and Stewart Slocum and Usman Anwar and Anand Siththaranjan and Max Nadeau and Eric J Michaud and Jacob Pfau and Dmitrii Krasheninnikov and Xin Chen and Lauro Langosco and Peter Hase and Erdem Biyik and Anca Dragan and David Krueger and Dorsa Sadigh and Dylan Hadfield-Menell},
journal={Transactions on Machine Learning Research},
issn={2835-8856},
year={2023},
url={https://openreview.net/forum?id=bx24KpJ4Eb},
note={Survey Certification, Featured Certification}
}

@misc{kaufmann2024surveyreinforcementlearninghuman,
      title={A Survey of Reinforcement Learning from Human Feedback}, 
      author={Timo Kaufmann and Paul Weng and Viktor Bengs and Eyke Hüllermeier},
      year={2024},
      eprint={2312.14925},
      archivePrefix={arXiv},
      primaryClass={cs.LG},
      url={https://arxiv.org/abs/2312.14925}, 
}

@misc{schulman2017proximalpolicyoptimizationalgorithms,
title={Proximal Policy Optimization Algorithms}, 
author={John Schulman and Filip Wolski and Prafulla Dhariwal and Alec Radford and Oleg Klimov},
year={2017},
eprint={1707.06347},
archivePrefix={arXiv},
primaryClass={cs.LG},
url={https://arxiv.org/abs/1707.06347}, 
}

@inproceedings{ouyang2022training,
author = {Ouyang, Long and Wu, Jeff and Jiang, Xu and Almeida, Diogo and Wainwright, Carroll L. and Mishkin, Pamela and Zhang, Chong and Agarwal, Sandhini and Slama, Katarina and Ray, Alex and Schulman, John and Hilton, Jacob and Kelton, Fraser and Miller, Luke and Simens, Maddie and Askell, Amanda and Welinder, Peter and Christiano, Paul and Leike, Jan and Lowe, Ryan},
title = {Training language models to follow instructions with human feedback},
year = {2022},
booktitle = {NeurIPS}
}

@inproceedings{dpo,
 author = {Rafailov, Rafael and Sharma, Archit and Mitchell, Eric and Manning, Christopher D and Ermon, Stefano and Finn, Chelsea},
 booktitle = {Advances in Neural Information Processing Systems},
 editor = {A. Oh and T. Naumann and A. Globerson and K. Saenko and M. Hardt and S. Levine},
 pages = {53728--53741},
 publisher = {Curran Associates, Inc.},
 title = {Direct Preference Optimization: Your Language Model is Secretly a Reward Model},
 url = {https://proceedings.neurips.cc/paper_files/paper/2023/file/a85b405ed65c6477a4fe8302b5e06ce7-Paper-Conference.pdf},
 volume = {36},
 year = {2023}
}

@misc{Bai2022ConstitutionalAH,
      title={Constitutional AI: Harmlessness from AI Feedback}, 
      author={Yuntao Bai and Saurav Kadavath and Sandipan Kundu and Amanda Askell and Jackson Kernion and Andy Jones and Anna Chen and Anna Goldie and Azalia Mirhoseini and Cameron McKinnon and Carol Chen and Catherine Olsson and Christopher Olah and Danny Hernandez and Dawn Drain and Deep Ganguli and Dustin Li and Eli Tran-Johnson and Ethan Perez and Jamie Kerr and Jared Mueller and Jeffrey Ladish and Joshua Landau and Kamal Ndousse and Kamile Lukosuite and Liane Lovitt and Michael Sellitto and Nelson Elhage and Nicholas Schiefer and Noemi Mercado and Nova DasSarma and Robert Lasenby and Robin Larson and Sam Ringer and Scott Johnston and Shauna Kravec and Sheer El Showk and Stanislav Fort and Tamera Lanham and Timothy Telleen-Lawton and Tom Conerly and Tom Henighan and Tristan Hume and Samuel R. Bowman and Zac Hatfield-Dodds and Ben Mann and Dario Amodei and Nicholas Joseph and Sam McCandlish and Tom Brown and Jared Kaplan},
      year={2022},
      eprint={2212.08073},
      archivePrefix={arXiv},
      primaryClass={cs.CL},
      url={https://arxiv.org/abs/2212.08073}, 
}

@book{mann1987rhetorical,
    title={Rhetorical structure theory: A theory of text organization},
    author={Mann, William C and Thompson, Sandra A},
    year={1987},
    publisher={University of Southern California, Information Sciences Institute Los Angeles}
}

@misc{ankner2024critiqueoutloudrewardmodels,
      title={Critique-out-Loud Reward Models}, 
      author={Zachary Ankner and Mansheej Paul and Brandon Cui and Jonathan D. Chang and Prithviraj Ammanabrolu},
      year={2024},
      eprint={2408.11791},
      archivePrefix={arXiv},
      primaryClass={cs.LG},
      url={https://arxiv.org/abs/2408.11791}, 
}

@inproceedings{ye2024improvingrewardmodelssynthetic,
      title={Improving Reward Models with Synthetic Critiques}, 
      author={Zihuiwen Ye and Fraser Greenlee-Scott and Max Bartolo and Phil Blunsom and Jon Ander Campos and Matthias Gallé},
      year={2025},
      booktitle={NAACL Findings},
      url={https://arxiv.org/abs/2405.20850}, 
}

@inproceedings{press2022trainshorttestlong,
      title={Train Short, Test Long: Attention with Linear Biases Enables Input Length Extrapolation}, 
      author={Ofir Press and Noah A. Smith and Mike Lewis},
      year={2022},
      booktitle={ICLR}
}

@inproceedings{dao2023flashattention2fasterattentionbetter,
      title={FlashAttention-2: Faster Attention with Better Parallelism and Work Partitioning}, 
      author={Tri Dao},
      year={2024},
      booktitle={ICLR}
}

@inproceedings{hu2021loralowrankadaptationlarge,
      title={LoRA: Low-Rank Adaptation of Large Language Models}, 
      author={Edward J. Hu and Yelong Shen and Phillip Wallis and Zeyuan Allen-Zhu and Yuanzhi Li and Shean Wang and Lu Wang and Weizhu Chen},
      year={2022},
      booktitle={ICLR}
}

@inproceedings{köksal2024longformeffectiveinstructiontuning,
      title={LongForm: Effective Instruction Tuning with Reverse Instructions}, 
      author={Abdullatif Köksal and Timo Schick and Anna Korhonen and Hinrich Schütze},
      year={2024},
      booktitle={EMNLP Findings}
}

@inproceedings{wu2025longgenbenchbenchmarkinglongformgeneration,
      title={LongGenBench: Benchmarking Long-Form Generation in Long Context LLMs}, 
      author={Yuhao Wu and Ming Shan Hee and Zhiqing Hu and Roy Ka-Wei Lee},
      year={2025},
      booktitle={ICLR}
}

@inproceedings{greenberg2021detecting,
  title={Detecting rewards deterioration in episodic reinforcement learning},
  author={Greenberg, Ido and Mannor, Shie},
  booktitle={International Conference on Machine Learning},
  pages={3842--3853},
  year={2021},
  organization={PMLR}
}

@inproceedings{tan2024proxyqaalternativeframeworkevaluating,
      title={PROXYQA: An Alternative Framework for Evaluating Long-Form Text Generation with Large Language Models}, 
      author={Haochen Tan and Zhijiang Guo and Zhan Shi and Lu Xu and Zhili Liu and Yunlong Feng and Xiaoguang Li and Yasheng Wang and Lifeng Shang and Qun Liu and Linqi Song},
      year={2024},
      booktitle={ACL}
}

@inproceedings{
zhang2024generative,
title={Generative Verifiers: Reward Modeling as Next-Token Prediction},
author={Lunjun Zhang and Arian Hosseini and Hritik Bansal and Mehran Kazemi and Aviral Kumar and Rishabh Agarwal},
booktitle={The 4th Workshop on Mathematical Reasoning and AI at NeurIPS'24},
year={2024},
url={https://openreview.net/forum?id=CxHRoTLmPX}
}

@inproceedings{chan2024denserewardfreereinforcement,
      title={Dense Reward for Free in Reinforcement Learning from Human Feedback}, 
      author={Alex J. Chan and Hao Sun and Samuel Holt and Mihaela van der Schaar},
      year={2024},
      booktitle={ICML}
}

@article{haugh2012speaker,
  title={Speaker intentions and intentionality},
  author={Haugh, Michael and Jaszczolt, Kasia M},
  journal={The Cambridge handbook of pragmatics},
  volume={87},
  pages={112},
  year={2012},
  publisher={Cambridge University Press Cambridge}
}

@article{degen2023rational,
  title={The rational speech act framework},
  author={Degen, Judith},
  journal={Annual Review of Linguistics},
  volume={9},
  number={1},
  pages={519--540},
  year={2023},
  publisher={Annual Reviews}
}

@article{pignatelli2024a,
title={A Survey of Temporal Credit Assignment in Deep Reinforcement Learning},
author={Eduardo Pignatelli and Johan Ferret and Matthieu Geist and Thomas Mesnard and Hado van Hasselt and Laura Toni},
journal={Transactions on Machine Learning Research},
issn={2835-8856},
year={2024},
url={https://openreview.net/forum?id=bNtr6SLgZf},
note={Survey Certification}
}

@misc{beltagy2020longformerlongdocumenttransformer,
      title={Longformer: The Long-Document Transformer}, 
      author={Iz Beltagy and Matthew E. Peters and Arman Cohan},
      year={2020},
      eprint={2004.05150},
      archivePrefix={arXiv},
      primaryClass={cs.CL},
      url={https://arxiv.org/abs/2004.05150}, 
}

@article{crossley2023large,
  title={A large-scale corpus for assessing written argumentation: PERSUADE 2.0},
  author={Crossley, Scott Andrew and Baffour, Perpetual and Tian, Yu and Franklin, Alex and Benner, Meg and Boser, Ulrich},
  journal={Available at SSRN 4795747},
  year={2023}
}

@inproceedings{10.1145/2872427.2883081,
author = {Tan, Chenhao and Niculae, Vlad and Danescu-Niculescu-Mizil, Cristian and Lee, Lillian},
title = {Winning Arguments: Interaction Dynamics and Persuasion Strategies in Good-faith Online Discussions},
year = {2016},
isbn = {9781450341431},
publisher = {International World Wide Web Conferences Steering Committee},
address = {Republic and Canton of Geneva, CHE},
url = {https://doi.org/10.1145/2872427.2883081},
doi = {10.1145/2872427.2883081},
booktitle = {Proceedings of the 25th International Conference on World Wide Web},
pages = {613–624},
numpages = {12},
location = {Montr\'{e}al, Qu\'{e}bec, Canada},
series = {WWW '16}
}

@misc{asap-aes,
    author = {Ben Hamner and Jaison Morgan and lynnvandev and Mark Shermis and Tom Vander Ark},
    title = {The Hewlett Foundation: Automated Essay Scoring},
    year = {2012},
    howpublished = {\url{https://kaggle.com/competitions/asap-aes}},
    note = {Kaggle}
}

@inproceedings{pathak2017curiositydrivenexplorationselfsupervisedprediction,
      title={Curiosity-driven Exploration by Self-supervised Prediction}, 
      author={Deepak Pathak and Pulkit Agrawal and Alexei A. Efros and Trevor Darrell},
      year={2017},
      booktitle={ICML}
}

@inproceedings{reddy2019sqilimitationlearningreinforcement,
      title={SQIL: Imitation Learning via Reinforcement Learning with Sparse Rewards}, 
      author={Siddharth Reddy and Anca D. Dragan and Sergey Levine},
      year={2019},
      booktitle={ICLR}
}

@inproceedings{NEURIPS2021_503e7dbb,
 author = {Chen, Jiayu and Zhang, Yuanxin and Xu, Yuanfan and Ma, Huimin and Yang, Huazhong and Song, Jiaming and Wang, Yu and Wu, Yi},
 booktitle = {Advances in Neural Information Processing Systems},
 editor = {M. Ranzato and A. Beygelzimer and Y. Dauphin and P.S. Liang and J. Wortman Vaughan},
 pages = {9681--9693},
 publisher = {Curran Associates, Inc.},
 title = {Variational Automatic Curriculum Learning for Sparse-Reward Cooperative Multi-Agent Problems},
 url = {https://proceedings.neurips.cc/paper_files/paper/2021/file/503e7dbbd6217b9a591f3322f39b5a6c-Paper.pdf},
 volume = {34},
 year = {2021}
}

@misc{yang2024qwen2technicalreport,
      title={Qwen2 Technical Report}, 
      author={An Yang and Baosong Yang and Binyuan Hui and Bo Zheng and Bowen Yu and Chang Zhou and Chengpeng Li and Chengyuan Li and Dayiheng Liu and Fei Huang and Guanting Dong and Haoran Wei and Huan Lin and Jialong Tang and Jialin Wang and Jian Yang and Jianhong Tu and Jianwei Zhang and Jianxin Ma and Jianxin Yang and Jin Xu and Jingren Zhou and Jinze Bai and Jinzheng He and Junyang Lin and Kai Dang and Keming Lu and Keqin Chen and Kexin Yang and Mei Li and Mingfeng Xue and Na Ni and Pei Zhang and Peng Wang and Ru Peng and Rui Men and Ruize Gao and Runji Lin and Shijie Wang and Shuai Bai and Sinan Tan and Tianhang Zhu and Tianhao Li and Tianyu Liu and Wenbin Ge and Xiaodong Deng and Xiaohuan Zhou and Xingzhang Ren and Xinyu Zhang and Xipin Wei and Xuancheng Ren and Xuejing Liu and Yang Fan and Yang Yao and Yichang Zhang and Yu Wan and Yunfei Chu and Yuqiong Liu and Zeyu Cui and Zhenru Zhang and Zhifang Guo and Zhihao Fan},
      year={2024},
      eprint={2407.10671},
      archivePrefix={arXiv},
      primaryClass={cs.CL},
      url={https://arxiv.org/abs/2407.10671}, 
}

@inproceedings{zheng2024sglangefficientexecutionstructured,
      title={SGLang: Efficient Execution of Structured Language Model Programs}, 
      author={Lianmin Zheng and Liangsheng Yin and Zhiqiang Xie and Chuyue Sun and Jeff Huang and Cody Hao Yu and Shiyi Cao and Christos Kozyrakis and Ion Stoica and Joseph E. Gonzalez and Clark Barrett and Ying Sheng},
      year={2024},
      booktitle={NeurIPS},
      url={https://arxiv.org/abs/2312.07104}, 
}

@inproceedings{lin2024awqactivationawareweightquantization,
      title={AWQ: Activation-aware Weight Quantization for LLM Compression and Acceleration}, 
      author={Ji Lin and Jiaming Tang and Haotian Tang and Shang Yang and Wei-Ming Chen and Wei-Chen Wang and Guangxuan Xiao and Xingyu Dang and Chuang Gan and Song Han},
      year={2024},
      booktitle={MLSys},
      url={https://arxiv.org/abs/2306.00978}, 
}

@misc{vonwerra2022trl,
  author = {Leandro von Werra and Younes Belkada and Lewis Tunstall and Edward Beeching and Tristan Thrush and Nathan Lambert and Shengyi Huang and Kashif Rasul and Quentin Gallouédec},
  title = {TRL: Transformer Reinforcement Learning},
  year = {2020},
  publisher = {GitHub},
  journal = {GitHub repository},
  howpublished = {\url{https://github.com/huggingface/trl}}
}

\end{document}